\documentclass[11pt]{article}
\usepackage[a4paper,bindingoffset=0.2in,%
            left=0.9in,right=0.9in,top=1in,bottom=1in,%
            footskip=.25in]{geometry}

\usepackage[utf8]{inputenc} 
\usepackage[T1]{fontenc}    
\usepackage{hyperref}       
\usepackage{url}            
\usepackage{booktabs}       
\usepackage{amsfonts}       
\usepackage{nicefrac}       
\usepackage{microtype}      

\usepackage{amsmath,amsthm}
\usepackage{graphicx}
\usepackage{latexsym}                
\usepackage{dsfont}                  
\usepackage{mathtools}               
\usepackage{xcolor}                  
\usepackage{subfigure}
\usepackage{pgf}
\usepackage{pgfpages}                

\usepackage[printonlyused]{acronym}
\usepackage{amssymb,amsmath}
\usepackage{bm}
\usepackage{apacite}

\newcommand{\real}{\mathbb{R}}

\renewcommand{\vec}[1]{\mathbf{\MakeLowercase{#1}}}
\newcommand{\mat}[1]{\MakeUppercase{#1}}
\newcommand{\set}[1]{\mathcal{#1}}

\newcommand{\T}{\text{T}}
\newcommand{\figref}[1]{Figure \ref{#1}}

\renewcommand{\epsilon}{\varepsilon}
\renewcommand{\cite}[1]{\shortcite{#1}}

\newtheorem{remark}{Remark}
\newtheorem{claim}{Claim}

\newtheorem{theorem}{Theorem}

\usepackage{algorithm,algorithmic}
\usepackage[colorinlistoftodos,prependcaption]{todonotes} 

\title{On the Effect of Low-Rank Weights on \\ Adversarial Robustness of Neural Networks}

\author{Peter~Langenberg,
		Emilio Rafael~Balda,
        Arash~Behboodi,
        and~Rudolf~Mathar
\thanks{Institute for Theoretical Information Technology (TI), RWTH Aachen University.}}

\begin{document}
\acrodef{DNN}{Deep Neural Network}
\acrodef{FGSM}{Fast Gradient Sign Method}
\acrodef{GNM}{gradient-based norm-constrained method}
\acrodef{BIM}{Basic Iterative Method}
\acrodef{PGD}{Projected Gradient Descent}
\acrodef{FCNN}{Fully Connected Neural Network}
\acrodef{CNN}{Convolutional Neural Network}

\maketitle

\begin{abstract}
Recently, there has been an abundance of works on designing \acp{DNN} that are robust to adversarial examples. 
In particular, a central question is which features of DNNs influence adversarial robustness and, therefore, can be to used to design robust \acp{DNN}. 
In this work, this problem is studied through the lens of compression which is captured by the low-rank structure of weight matrices.
It is first shown that adversarial training tends to promote simultaneously low-rank and sparse structure in the weight matrices of neural networks.
This is measured through the notions of effective rank and effective sparsity.
In the reverse direction, when the low rank structure is promoted by nuclear norm regularization and combined with sparsity inducing regularizations, neural networks show significantly improved adversarial robustness.
The effect of nuclear norm regularization on adversarial robustness is paramount when it is applied to convolutional neural networks.  Although still not competing with adversarial training, this result contributes to understanding the key properties of robust classifiers. 
%
%
%
%
%
\end{abstract}

\section{Introduction}
\label{sec:intro}
In recent years, deep neural networks (DNNs) have greatly improved the state-of-the-art performance of a variety of tasks in computer vision 
\cite{HeZR015} 
and automatic speech recognition \cite{XiongDHSSSYZ16a}, medical imaging \cite{LitjensKBSCGLGS17} \cite{ esteva2017dermatologistlevel}, robotics \cite{limitsRobotics18} \cite{giusti2016machine} and game playing \cite{silver2017mastering}.
However, it has been shown that very small additive perturbations of the input sample, known as \textit{adversarial} perturbations, can cause DNN based vision systems to fail \cite{intriguing} \cite{harness_nnn}. 
Such intentionally perturbed versions of clean samples that fool DNNs  
are referred to as \textit{adversarial examples}.
The robustness of DNNs against these attacks has recently received significant attention. 

Different hypotheses have been made about the existence of adversarial examples. 
In \cite{harness_nnn}, it is hypothesized that DNNs are particularly vulnerable to adversarial examples because of their too linear decision boundary in the vicinity around the data points together with the assumption of sufficiently large dimensionality of the problem. 
In \cite{TanayG16}, however, it is shown that it is possible to train linear classifiers that are resistant to adversarial attacks which stands in contrast to the linearity hypothesis. 
Moreover, it is exemplified that high dimensional problems are not necessarily more sensitive to adversarial examples. 
Further, \cite{sabour2015adversarial} 
manipulated deep representations instead of the input and argued that the 
linearity hypothesis is not sufficient to explain this type of attack. 
Regarding the observed transferability of adversarial examples across different DNN architectures \cite{intriguing} and \cite{Tramr2017TheSO}
propose a method for estimating the dimensionality of the space of adversarial examples. It is shown that adversarial examples
span a contiguous subspace of large dimensionality. Further, it is shown that those subspaces intersect for different DNNs that have common adversarial examples.    
\cite{flexi} propose the low flexibility of DNNs, compared to the difficulty of the classification task, as a reason for the existence of adversarial examples. 
\cite{NIPS2016_6331} suggest that the flatness of the decision boundary is a reason for the existence of adversarial examples. 
Another perspective is proposed in \cite{TanayG16}
with the \textit{boundary tilting} mechanism. It is argued that
adversarial examples exist when the
decision boundary lies close to the sub-manifold of sampled data. 
The notion of \textit{adversarial strength} is introduced which refers to the deviation angle
between the target classifier and the nearest centroid classifier. 
It is shown that the adversarial strength can be arbitrarily increased independently of the classifier's accuracy by tilting the boundary.  
\cite{RozsaGB16b} give another explanation arguing that over the course of training 
the correctly classified samples do not have a significant impact on shaping the decision boundary and eventually remain close to it. This phenomenon is called evolutionary stalling.
In \cite{Rozsa2016AreAA}, the correlation between robustness and accuracy of the classifier is studied empirically by attacking different state-of-the-art DNNs. It is observed that higher accuracy DNNs are more sensitive to adversarial attacks than lower accuracy ones. 
Regarding universal adversarial examples, \cite{MoosaviDezfooli2017UniversalAP} show empirically and \cite{moosavi-dezfooli2018robustness} formally that their existence is partly caused by the correlation between the normals to the decision boundary in the vicinity of natural images, i.e. these normals span a low dimensional space. It is found that in this subspace the decision boundary is positively curved in the vicinity of natural images. 
In that work, \cite{moosavi-dezfooli2018robustness} assume that the first order linear approximation of the decision boundary holds locally in the vicinity of natural images. 
However, \cite{moosavi-dezfooli2018robustness} show also that this model is not sufficient to explain large fooling ratios. The authors introduce the {locally curved decision boundary model} 
which indicates that 
DNNs are particularly vulnerable to universal perturbations in shared directions  
along which the decision boundary is systematically positively curved, i.e. there exists a subspace where the decision boundary is positively curved in the vicinity of natural images along most directions in this subspace.

In \cite{scale} and \cite{madry2018towards}, evidence was given that increasing model capacity alone can help to induce \acp{DNN} to be more robust against adversarial attacks. Further, it was observed that robust models (obtained by adversarial training) exhibit rather sparse weights compared to unrobust ones. 
\cite{madry2018towards} compare the weights of a CNN that is trained naturally vs. adversarially on the MNIST dataset \cite{lecun-mnisthandwrittendigit-2010}. After training, it is observed that the adversarially trained CNN has more sparse weights in the first two convolutional filters  than the naturally trained one. \cite{gopalakrishnan2018combating, sparsity_defense} propose that producing sparse input representations improves the robustness of linear and deeper binary classifiers. \cite{sparse_dnns} conduct a more detailed study of the relation between sparsity and robustness of DNNs. It is shown formally and empirically that more sparse DNNs are more robust against adversarial attacks, up to a certain limit where robustness starts decreasing again.  
Recently, \cite{low_rank_repr} have provided evidence that adversarial robustness can be improved by encouraging the learned representations to lie in a low-rank subspace.
We see that there are many different hypotheses about the nature of adversarial examples which, in fact, do not perfectly match. This makes the research on the nature of adversarial examples an active area.
%
%
%
Inspired by recent works \cite{madry2018towards, sparse_dnns, low_rank_repr} in this work we suggest that one should design robust DNNs considering the \textit{effective sparsity} and the \textit{effective rank} of the weights. 
We observe that adversarial training leads to low-rank and sparse weights. So far, these two properties have not been pointed out together in the context of adversarial robustness of DNNs. 
We are able to substantially improve adversarial robustness by simultaneously promoting low-rank and sparse weights using different regularization techniques. 
Moreover, we raise an open question whether there is an optimal combination of sparsity and low-rankness of the weights that can further improve robustness. 
We believe that simultaneous sparsity and low-rankness of the weights play a significant role in robustness, but do not fully explain the success of adversarial training. 

\section{Preliminaries}
\label{sec:prelimi}
Let $\vec{x} \in \set{X} \subseteq \mathbb{R}^n$ to be a random $n$-dimensional vector representing the input of classifiers, which belongs to some set of possible inputs $\set{X}$.
Furthermore, we denote $\hat{\vec{y}} \in \set{Y}$ to be the output label assigned by a classifier,
which belongs to some discrete set $\set{Y}$ of possible labels.
Then a classifier consists of some function $f:\Theta \times \set{X} \rightarrow \set{Y}$ such that $\hat{\vec{y}} = f(\vec{\theta}; \vec{x})$ where $\vec{\theta} \in {\Theta}$ contains all tunable model parameters in the hypothesis space ${\Theta}$.
In the case of \ac{DNN} of $m$ layers the set of tunable parameters is given by $\theta = \{W^1, \dots, W^m\}$ where $W^i$ are matrices with appropriate dimensions such that the neural network function $f$ has the form
\[
f(\vec{\theta}; \vec{x}) = \varphi( \mat W^m \varphi( \mat W^{m-1} ( \cdots \varphi( \mat W^1 \vec{x}  )) ) \cdots) \, ,
\]
where $\varphi:\mathbb{R} \rightarrow \mathbb{R}$ is an activation function that is applied to vectors in a point-wise manner.
For the sake of notation let $i \in [m]$ be the layer index, $m$ the number of hidden and output layers,
and $m^i$ the number of neurons of layer $i$.
The vectors $\vec{t}^i \in \mathbb{R}^{m^i}~ \forall i \in \{1, \dots, m-1 \}$ are called hidden representations, given by
\begin{equation} \label{eq:deep_compact}
 \vec{t}^i = \varphi (\mat W^i \vec{t}^{i-1}) \, , \quad \forall \, i=1,\dots,m \, ,
\end{equation}
with $\vec{t}^0 = \vec{x}$ and $\hat{\vec{y}} = \vec{t}^m$.
Each layer is represented by its own weight matrix $\mat W^i \in \mathbb{R}^{m^i \times m^{i-1}}$ that defines the mapping from $\vec{t}^{i-1}$ to $\vec{t}^{i}$.
In this work, we refer to \acp{FCNN} to the \acp{DNN} that only contain fully-connected layers, while \acp{CNN} are \acp{DNN} containing convolutional filters.
We use the operators $\|\cdot \|_{\text{F}}$, $\|\cdot \|_{*}$ and $\|\cdot \|_{1}$ to denote the Frobenius, nuclear and $\ell_1$ norms respectively.
%
%
%
The effective sparsity of a
matrix $\mat W$ is given by
\begin{equation} \label{eq:esparse}
\overline s (\mat W)
=  \frac{\| \mat W \|_{1}}{\|\mat W\|_{\text{F}}} \, .
\end{equation}
The definition applies to vectors as well with $\overline s (\vec x)=  \frac{\| \vec x \|_{1}}{\|\vec x\|_2}$. Note that if a vector $\vec{x}$ is $s$-sparse, i.e., has only $s$ non-zero values, we have $\|\vec x\|_1\leq \sqrt s\|x\|_2$. In that sense, this notion is commonly used in compressed sensing as an extension of standard sparsity, see \cite{Gopi_effective_sparsity}.
Similarly, the effective sparsity of the vector of singular values of a matrix $\mat W$ is called the effective rank of such matrix, that is
\begin{equation} \label{eq:erank}
\overline r (\mat W)
=  \frac{\| \mat W \|_{*}}{||\mat W \|_{\text{F}}} \, .
\end{equation}
which is a continuous relaxation of the notion of rank.

Studying \acp{DNN} by measuring information theoretic quantities during their training phase 
has been initiated by \cite{Tishby2015DeepLA, Shwartz-ZivT17}.
The authors propose that the hidden representations $\vec{t}^i$ of a DNN form the following Markov Chain
\begin{equation}
    (\vec{x}, \vec{y}) \rightarrow \vec{t}^1 \rightarrow ... \rightarrow \vec{t}^{m-1} \rightarrow \hat{\vec{y}} \, .
\end{equation}
Such successive representations are studied with the notion of \textit{mutual information},
which quantifies how much information about one random variable can be obtained if the other random variable is observed.
We use the notation $I(\vec{a} ; \vec{b} )$ for the mutual information between two random variables $\vec{a}, \vec{b} $.
Using this quantity, the authors study the behavior of
the so called \textit{information plane} which is the $2$-dimensional plane of  values between successive representations and the input $I(\vec{t}^i; \vec{x})$ and the output $I(\vec{t}^i; \vec{y})$ of \acp{DNN} during training.
In this context, the hidden representation $\vec{t}^i$ is seen as a compressed version of $\vec{x}$, where information is lost during compression.
Then, $I(\vec{t}^i; \vec{x})$ quantifies the amount of information about $\vec x$ that is contained in $\vec{t}^i$.
Even though this paper focuses on the low-rankness and sparsity of the weights to quantify compression, we show experimentally that our findings align with the information theoretic notion of compression, which is measured using mutual information.
\section{Simultaneous Low-Rankness and Sparsity in Linear Models: An Example} 
\label{sec:simultaneous_low_rankness_and_sparsity_in_linear_models}
Consider the case of linear binary classifiers as an exmaple. Let $y \in  \{-1, +1\}$ be the random variable corresponding to the label of an input $\vec{x}\in\mathbb{R}^n$. 
The linear classifier with parameter $\vec{w}\in\real^n$ assigns  the label  $\hat y = \mathrm{sign}( \vec{w}^\T \vec{x} )$ to an input $\vec{x}$.  
Note that we neglect the bias term $b$ as in $\mathrm{sign}( \vec{w}^\T \vec{x} + b)$ since it can be easily included by appending a scalar $1$ to the input vector $\vec x$.
Therefore, the expected error of this classifier, denoted by $p_0(\vec w)$, is given by
\[
p_0(\vec w) = P(\vec{w}^\T \vec{x} \cdot y < 0 ) \, .
\]
An $\ell_\infty$ adversarial attack against this classifier is obtained by an input perturbations $\bm \eta$ whose $\ell_\infty$-norm is bounded by some $\epsilon > 0$.
The adversarial error is characterized by the following probability: 
\[
p_\epsilon(\vec{w}) = \max_{\bm \eta : \|\bm \eta\|_\infty \leq \epsilon} P(\vec{w}^\T (\vec{x} + \bm \eta)\cdot y < 0) \, .
\]
The probability $p_\epsilon(\vec{w})$ is a measure of robustness for linear classifiers. The goal is to minimize the error by an appropriate choice of $\vec{w}$.
A typical choice of $\bm\eta$ is given by the so called \ac{FGSM} attack for which $\bm\eta=-y\epsilon\mathrm{sign}(\vec{w})$. It was previously discussed in  \cite{sparse_dnns} that the robustness of linear classifiers are upper bounded by  the inverse of $\|\vec{w}\|_1$. Since the $\ell_1$-norm is widely used as a sparsity promoting regularization, the authors concluded that the sparsity of $\vec{w}$ contributes to the robustness of classifiers against $\ell_\infty$ attacks. As a matter of fact it can be seen that the robustness to attacks with a bounded norm is related to small dual norm of $\vec{w}$. In this work, we show that the robustness can be improved if the sparsity is complemented with an adequate dimensionality reduction. Although the example of binary linear classifiers is only considered, the result can be also extended to the multi-class setup similar to \cite{sparse_dnns}.

In most cases, high dimensional data belong to a low dimensional manifold. In this section, we assume that the data lie in a $d$-dimensional subspace $\set{V} \subset\real^n$ spanned by $d$ orthonormal basis $\vec{v}_1,\dots,\vec{v}_k$ with $k<n$. However, since the distribution of data is unknown, the dimensionality of data is not known. Therefore,
we apply an intermediate linear transformation $\mat{Q}$ on the data and try to reduce the dimensionality based on the training samples. The classifier is then applied to $\mat Q\vec x$ for $\vec{x}\in\real^n$. The main question is whether this compression step affects adversarial robustness. Intuitively an $\ell_\infty$-norm attack allows for $\epsilon$ perturbation of each entry which can translate each data point at by the distance of $\epsilon\sqrt{n}$ in the space. For higher dimension $n$,  $\ell_\infty$-norm attacks can create a bigger overall perturbation of the points. Therefore reducing the effective dimension seems to be favorable to adversarial robustness. A first choice of $\mat{Q}\in\real^{n\times n}$ can be the orthogonal projection matrix onto the data subspace $\set{V}$ defined by
\[
\mat{Q}=\sum_{i=1}^k \vec{v}_i\vec{v}_i^T.
\]
The following theorem shows that under certain assumptions, this projection can indeed improve the robustness even further.

%
%
\begin{theorem}\label{theo:robustness}
For a binary classification task, suppose that the data samples belong to $k$-dimensional subspace $\set{V}$ and $\mat{Q} \in\real^{n\times n}$ is the orthogonal projection onto this subspace.  Then, for any given binary linear classifiers with the parameter  $\vec{w}$,  if the effective sparsity of $\vec{w}$ after projection satisfies $\overline s(\mat Q\vec{w}) \leq \overline s(\vec{w})$, then
\begin{itemize}
	\item The accuracy remains unchanged, \textit{i.e.},  
\[
p_0(\vec w)=p_0(\mat{Q}{\vec w}).
\]
\item The adversarial robustness against $\ell_\infty$-attacks is either improved or unchanged, \textit{i.e.},
\[
p_\epsilon(\mat Q {\vec{w}})
\leq
p_\epsilon(\vec{w}).
\]  	 
\end{itemize}

%
\end{theorem}
\begin{remark}
	Before stating the proof, some remarks are in order. First of all, 	note that the new classifier applies to the projected vectors $\mat{Q}\vec{x}$. Since $\mat{Q}$ is symmetric, this amounts to a new binary classifier with parameter $\mat{Q}{\vec{w}}$. As it will be seen from the proof, if the vector $\vec{w}$ already belongs to the data subspace $\set{V}$, that is, the discriminating hyperplane is orthogonal to $\set{V}$, then the adversarial robustness remains unchanged. However, this is highly unlikely for most datasets that are noisy and therefore difficult to find exactly this hyperplane among many choices.
\end{remark}
\begin{proof}
The optimal $\ell_\infty$ attack for a vector $\vec x$ with label $y$ is given by $\bm\eta=-\epsilon y\mathrm{sign}(\vec{w})$. The adversarial robustness is therefore characterized by
\begin{align*}
	&P(\vec{w}^T(\vec x+\bm\eta).y>0)=P(y.\vec{w}^T\vec{x}>\epsilon \|\vec{w}\|_1)\\
	&=(1-p_0(\vec{w})) P(|\vec{w}^T\vec{x}|>\epsilon \|\vec{w}\|_1\mid y.\vec{w}^T\vec{x}>0)
\end{align*}
First suppose that the same classifier is applied after the orthogonal projection. Note that for all $\vec x\in\set{V}$, $\mat{Q}\vec x=\vec x$ and since the data belongs to $\set{V}$, we have
$p_0(\vec{w})=p_0(\mat{Q}\vec{w})$. Furthermore, if  $\bm\eta$ is chosen according to \ac{FGSM}, we get:
\begin{align*}
	&P(\vec{w}^T\mat{Q}(\vec x+\bm\eta).y>0)=P(y.\vec{w}^T\mat{Q}\vec{x}>\epsilon \|\mat{Q}\vec{w}\|_1)\\
	&=(1-p_0(\vec{w})) P(|\vec{w}^T\vec{x}|>\epsilon \|\mat{Q}\vec{w}\|_1\mid y.\vec{w}^T\vec{x}>0).
\end{align*}
If $\|\mat{Q}\vec{w}\|_1\leq \|\vec{w}\|_1$, the theorem follows.  Since $\mat Q$ is a projection matrix, for any $\vec w$ we have that $ \| \mat Q \vec w \|_2 \leq \| \vec w \|_2$. This inequality and the definition of effective sparsity yield
	\[
	 \frac{\|\mat Q \vec w\|_1}{ \overline s(\mat Q \vec w)} \leq \frac{\|\vec w\|_1}{ \overline s(\vec w)} \, .
	\]
	Given the initial assumption $\overline s(\mat Q\vec{w}) \geq \overline s(\vec{w})$ we get 
	\[
	 \|\mat Q \vec w\|_1 \leq \frac{\overline s(\mat Q \vec w)}{\overline s(\vec w)} \|\vec w\|_1 \leq \|\vec w\|_1\, .
	\] 
	Therefore $\epsilon \|\vec{w}\|_1 \geq \epsilon \|\mat Q\vec{w}\|_1$ thus $p_\epsilon(\vec{w}) \geq p_\epsilon(\mat Q \vec{w})$.
\end{proof}
%
%
 The condition on effective sparsity is not very demanding. Indeed assume that $\set{V}=\mathrm{span}(\vec e_1,\dots,\vec e_k)$ for $k<n$. We have $\|\mat{Q}\vec{w}\|=\sum_{i=1}^k |w_i|\leq \|\vec w\|_1$. The equality holds only if $\vec{w}\in\set{V}$. In other words, if the discriminating hyperplane is not orthogonal to the data subspace, there is always a gain in low dimensional projection. Note that for an arbitrary classifier $f(\cdot)$, the accuracy remains unchanged after the projection. In this case, if the corresponding gradient is projected such that its $\ell_1$ norm is reduced after the projection, then the robustness will be increased. Finally, if we overparametrize the classifier $\mathrm{sign}((\mat Q \vec w)^\T \vec x)$ as $\mathrm{sign}((\mat Q \mat W^1 \cdots \mat W^{d-1} \vec w)^\T \vec x)$ we obtain a $d$-layered neural network with linear activations whose initial weight matrix given by $\mat Q \mat W^1$ has low rank. This remark points our attention specially into the low-rankness the first weight matrices.
\section{Inducing Compression through Regularization} 
\label{sec:inducing_compression_through_regularization}
\begin{figure}[tbh]
    \centering
    \begin{minipage}[b]{0.31\linewidth}
\includegraphics[width=0.98\linewidth]{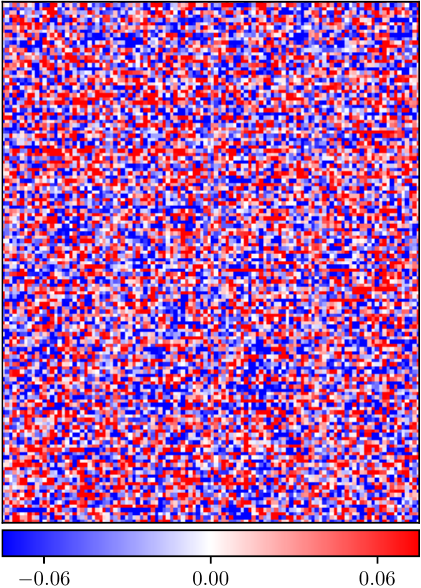}
            \centerline{\footnotesize{\textbf{(a)} Natural training}}\medskip
    \end{minipage} %
    \begin{minipage}[b]{0.31\linewidth}
\includegraphics[width=0.98\linewidth]{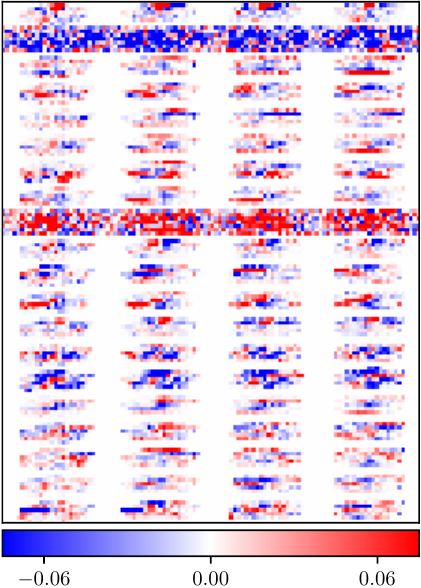}
            \centerline{\footnotesize{\textbf{(b)} \ac{FGSM} training}}\medskip
    \end{minipage} %
     \begin{minipage}[b]{0.31\linewidth}
\includegraphics[width=0.98\linewidth]{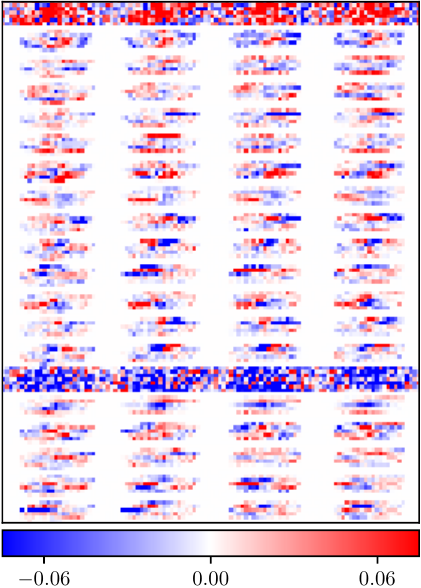}
            \centerline{\footnotesize{\textbf{(c)} \ac{PGD} training}}\medskip
    \end{minipage} %
    \caption{Reshaped input weight matrix $\mat{W}^1 \in \mathbb{R}^{20 \times 784}$ of the \ac{FCNN} after natural vs. adversarial training with $\epsilon = 0.05$.}
    \label{fig:first_fc_kernel_adv_train}
\end{figure}

\begin{figure*}[tbh]
   \centering
   \includegraphics[width=\linewidth]{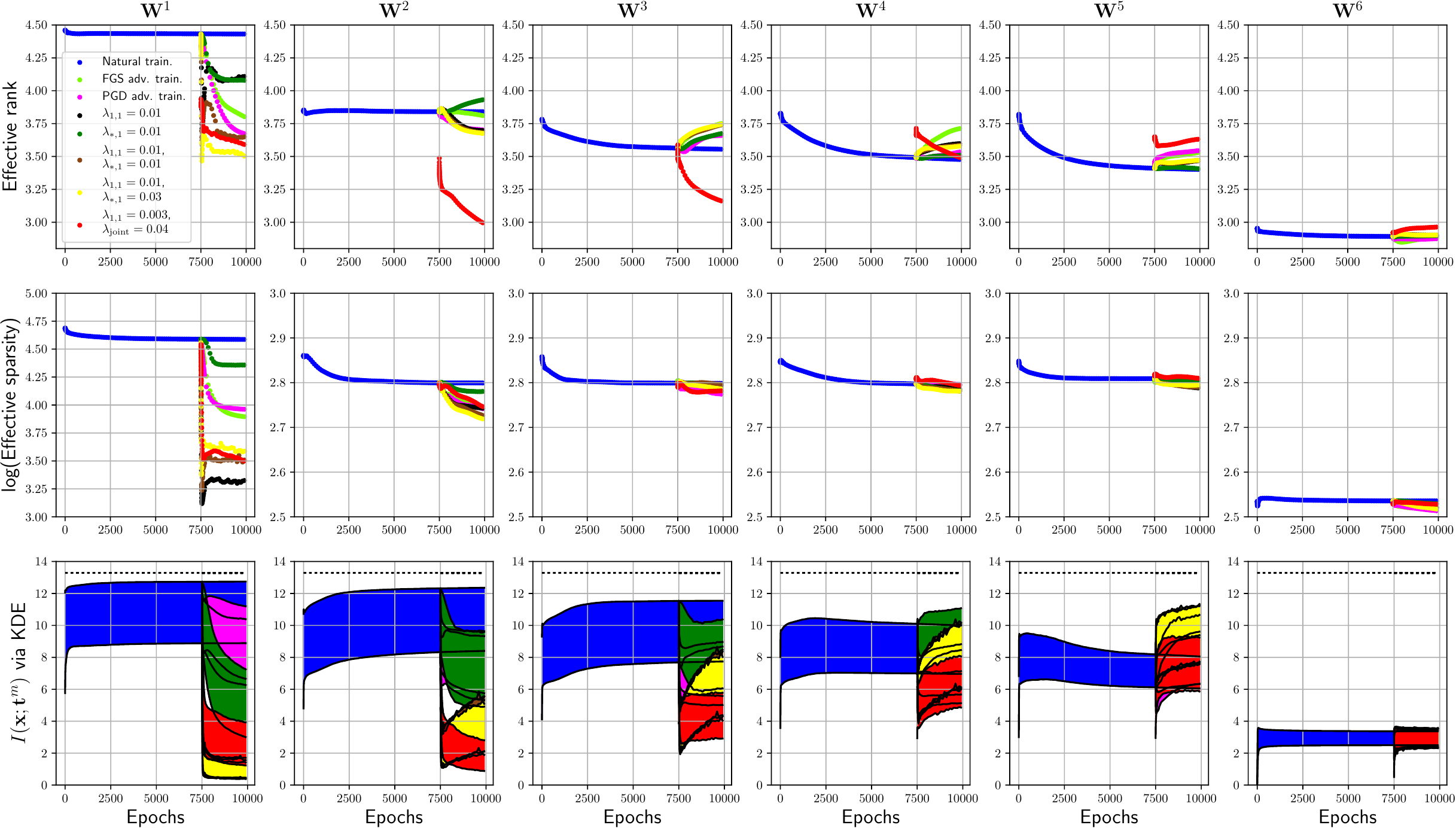}
   \caption{Effective rank, effective sparsity and mutual information of the FCNN. }
   \label{fig:compression_fc}
\end{figure*}

%
%
In \figref{fig:first_fc_kernel_adv_train} and \figref{fig:compression_fc}, we observe how adversarial training, using the \ac{FGSM} and \ac{PGD} attacks, induces effective sparsity as well as effective low-rankness to the weight matrices of a \ac{FCNN} (details about the simulation setup are provided in section \ref{sec:experiments}). Interestingly, the effective low-rankness of weights coincide with  lower mutual information values between the input and hidden representations which can be seen as a form of compression in the information theoretic sense.
%
%
%
%
%
Motivated by this result,  as well as Theorem \ref{theo:robustness}, the question is whether it is possible to improve the adversarial robustness by promoting sparsity and low-rankness without adversarial training.
We introduce three regularization techniques designed to promote sparsity and low-rankness to the weight matrices of \acp{DNN}.
In order to promote sparsity on the weight matrices, the $\ell_1$-regularization, a common choice for this purpose, is employed:
\begin{equation}
    \mathcal{L}_{\ell_1}(\vec{\theta}, \bm{\lambda}_1; \vec{x}, \vec{y}) =
\sum_{i=1}^{m} \lambda_{1,i} \| \mat W^i \|_{1} \, ,
\end{equation}
%
where $\bm{\lambda}_1 = (\lambda_{1,1} ,\dots , \lambda_{1,m}) \in \real^m_+$ is a vector of hyper-parameters. Just like $\ell_1-$regularization is a relaxation of sparsity measure, the nuclear norm is the convex relaxation of rank of a matrix and therefore,  the nuclear norm regularization is used to promote low-rankness.
\begin{equation} \label{eq:nuc_reg}
\mathcal{L}_{\text{nuclear}}(\vec{\theta}, \bm{\lambda}_*; \vec{x}, \vec{y}) =
\sum_{i=1}^{m} \lambda_{*,i} \| \mat W^i \|_{*} \, ,
\end{equation}
with the corresponding hyper-parameters $\bm{\lambda}_* = (\lambda_{*,1} ,\dots , \lambda_{*,m}) \in \real^m_+$.
%
%

%

As discussed in Section \ref{sec:simultaneous_low_rankness_and_sparsity_in_linear_models} and in \cite{sparse_dnns} the robustness to $\ell_\infty$ attacks of multilayer neural networks with linear activations is controlled by the $\ell_1$-norm of the product of its weight matrices, that is $\| \mat W^1 \cdots \mat W^m \|_{1}$.
Motivated by this result,
we include into our study the following regularization
\begin{equation} \label{eq:joint}
    \mathcal{L}_{\text{joint}}(\vec{\theta}, \lambda_{\text{joint}}; \vec{x}, \vec{y}) =
\lambda_{\text{joint}} \| \mat W^1 \cdots \mat W^m \|_{1}
\end{equation}
which promotes robustness in linear neural networks. It might lead to robustness for non-linear networks if the linear model is an adequate approximation.
Note that $\mathcal{L}_{\text{joint}}(\vec{\theta}, \lambda_{\text{joint}}; \vec{x}, \vec{y})$ is cheaper to compute than $\mathcal{L}_{\text{nuclear}}(\vec{\theta}, \vec{\lambda}_*; \vec{x}, \vec{y})$, which makes it a good alternative to avoid computing SVDs during training. Finally, incorporating these regularization losses into one global loss function yields
\begin{align}
    \mathcal{L}(\vec{\theta}; \vec{x}, \vec{y}) =& \mathcal{L}_{\text{CE}}(\vec{\theta}; \vec{x}, \vec{y}) + \mathcal{L}_{\ell_1}(\vec{\theta}, \bm{\lambda}_1; \vec{x}, \vec{y})
    \\
    & + \mathcal{L}_{\text{nuclear}}(\vec{\theta}, \bm{\lambda}_*; \vec{x}, \vec{y}) + \mathcal{L}_{\text{joint}}(\vec{\theta}, \lambda_{\text{joint}}; \vec{x}, \vec{y}) \, .
    \nonumber
\end{align}
%
%
%
In \figref{fig:compression_fc}, we observe how the proposed regularization techniques successfully induce low-rankness and sparsity to the weight matrices. At the same time, the mutual information between the input and hidden representations is reduced.

\section{Experiments} 
\label{sec:experiments}

In this work, we consider two architectures and datasets: 
\begin{enumerate}
\item \textbf{\ac{FCNN}}: 
is a fully-connected neural network with 5 hidden layers used on the MNIST handwritten digits dataset \cite{lecun-mnisthandwrittendigit-2010}. Each hidden layer contains $20$ neurons with hyperbolic tangent activations, while the output layer uses the $\mathrm{softmax}$ function. We use a learning rate of $0.001$ and a batch size of $128$ for both training from scratch and fine-tuning. 
\item \textbf{\ac{CNN}}: 
is a convolutional neural network applied on the Fashion-MNIST (F-MNIST) clothing articles dataset \cite{f-mnist}. 
It is composed two convolution/max-pooling blocks followed by two fully connected layers.
Both convolution/max-pooling blocks use $32$ convolution filters of window size $3 \times 3$ and pooling window of size $2 \times 2$. 
The first fully connected layer outputs $128$ outputs with hyperbolic tangent activations, while the output layer outputs $10$ $\mathrm{softmax}$ values.
We use a batch size of $64$ with a learning rate of $0.01$ for training from scratch, and $0.001$ for fine-tuning.  
\end{enumerate}
In both cases stochastic gradient descent is used for minimizing the loss function.

We consider the \ac{FGSM} \cite{harness_nnn}, \ac{PGD} \cite{madry2018towards} and the \ac{GNM} from \cite{balda2018adversarialconvex} as adversarial attacks and study their success on DNNs that are trained naturally, adversarially with \ac{FGSM}/\ac{PGD} adversarial examples as well as when using the aforementioned regularization techniques (see section \ref{sec:inducing_compression_through_regularization}). 
The fooling ratio is defined as the percentage of inputs, among the correctly classified ones, whose classification outcome changes after adding adversarial perturbations. 
This ratio is calculated on the test set as an empirical measure of sensitivity of the classifier against adversarial attacks for a given upper bound $\epsilon$ on the adversarial perturbation.
We focus on the low $\epsilon$ regime since small $\epsilon$ corresponds to adversarial examples that are hard to detect by an observer. 
In such regime the \ac{FGSM} often leads to similar attacks as \ac{PGD} since the decision boundary is linear enough in a vicinity around the clean sample. 
We found that $\epsilon = 0.05$ was a proper choice that leads to a fooling ratio 
slightly below $100 \%$ in case of natural training for both considered architectures.
We generate the attack using the \ac{GNM} from \cite{balda2018adversarialconvex} with $T=5$ iterations, while 
PGD attacks are generated with $T=20$ iterations and a step size of $\alpha = 0.005$.

\begin{figure*}[tbh]
	\centering
    \begin{minipage}[b]{0.31\linewidth}
        	\includegraphics[width=0.98\linewidth]{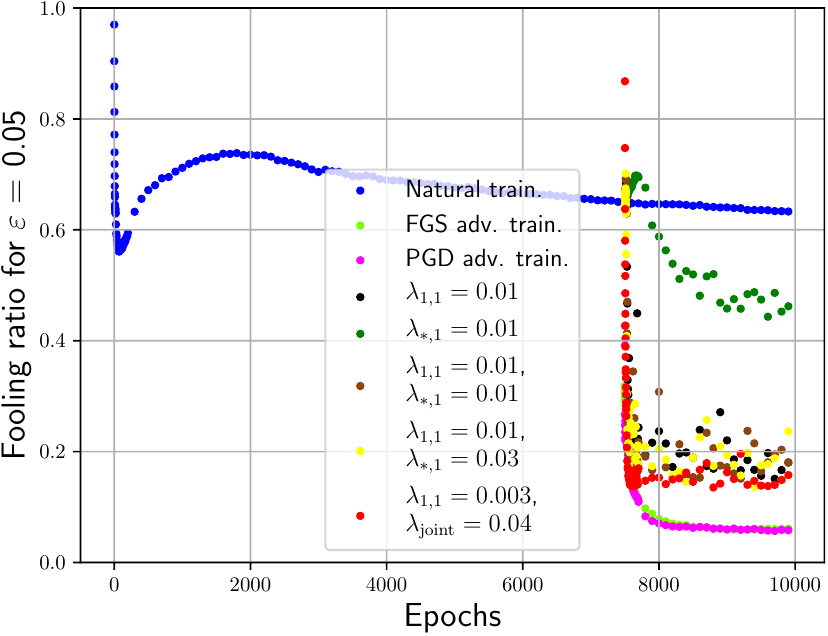}
            \centerline{\footnotesize{\textbf{(a)} \ac{FGSM}}}\medskip
	\end{minipage} %
    \begin{minipage}[b]{0.31\linewidth}
        	\includegraphics[width=0.98\linewidth]{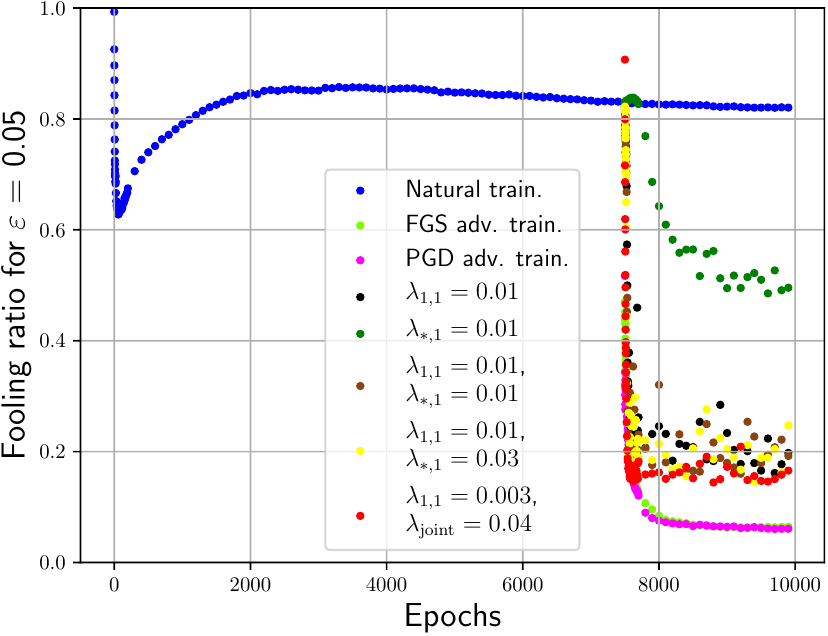}
            \centerline{\footnotesize{\textbf{(b)} \ac{GNM}}}\medskip
	\end{minipage} %
	 \begin{minipage}[b]{0.31\linewidth}
        	\includegraphics[width=0.98\linewidth]{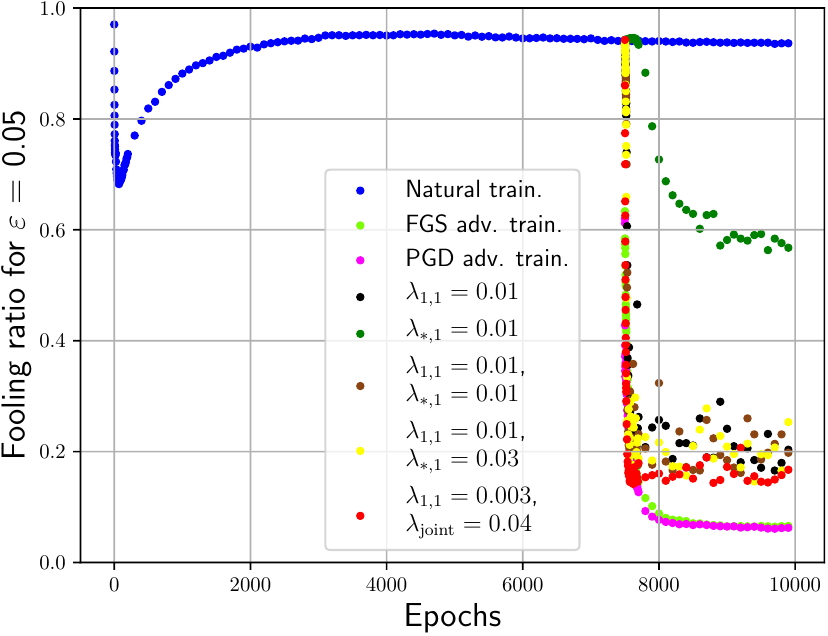}
            \centerline{\footnotesize{\textbf{(c)} \ac{PGD}}}\medskip
	\end{minipage} %
    \caption{Fooling ratios of the \ac{FCNN} with respective attack and defense using a fixed $\epsilon = 0.05$. Dataset: MNIST.}
\label{fig:fool_fc}    
\end{figure*}

\begin{figure*}[tbh]
	\centering
    \begin{minipage}[b]{0.31\linewidth}
        	\includegraphics[width=0.98\linewidth]{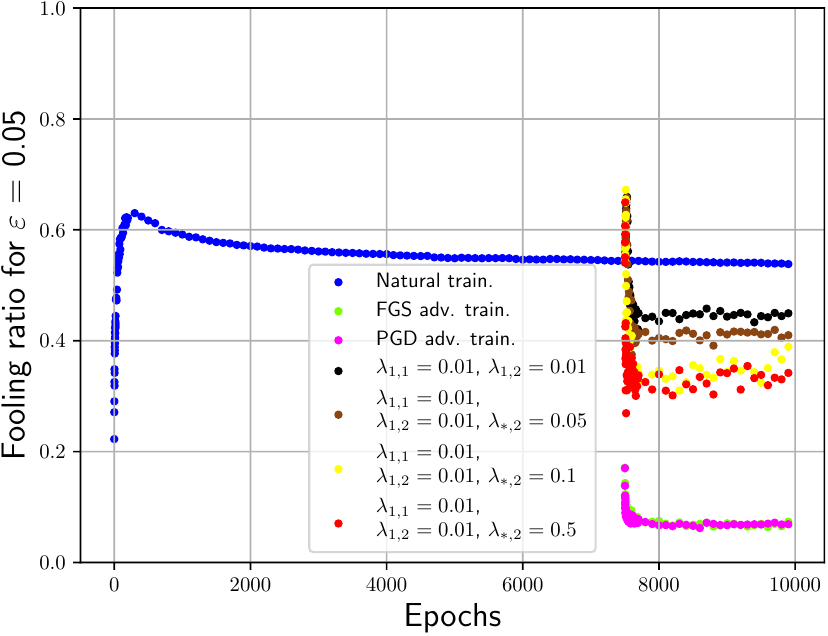}
            \centerline{\footnotesize{\textbf{(a)} \ac{FGSM}}}\medskip
	\end{minipage} %
    \begin{minipage}[b]{0.31\linewidth}
        	\includegraphics[width=0.98\linewidth]{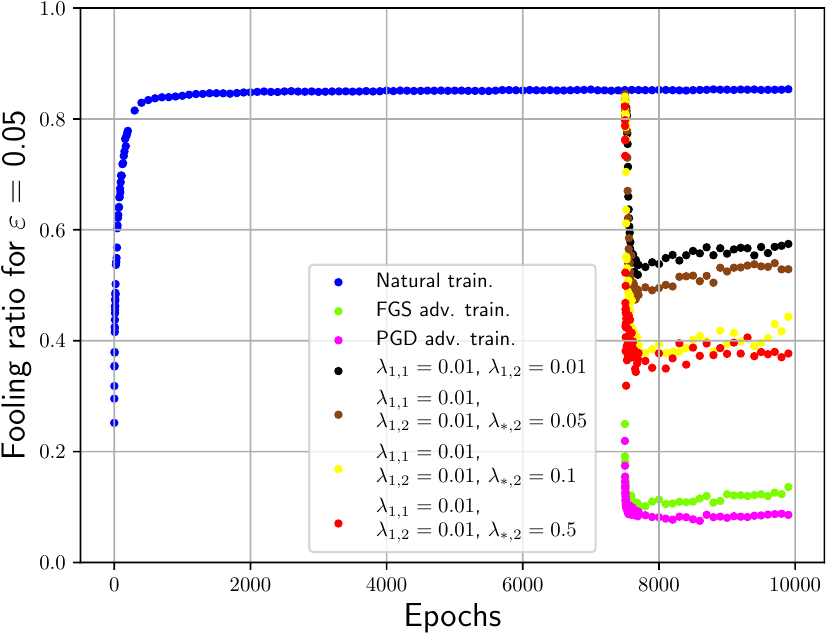}
            \centerline{\footnotesize{\textbf{(b)} \ac{GNM}}}\medskip
	\end{minipage} %
	 \begin{minipage}[b]{0.31\linewidth}
        	\includegraphics[width=0.98\linewidth]{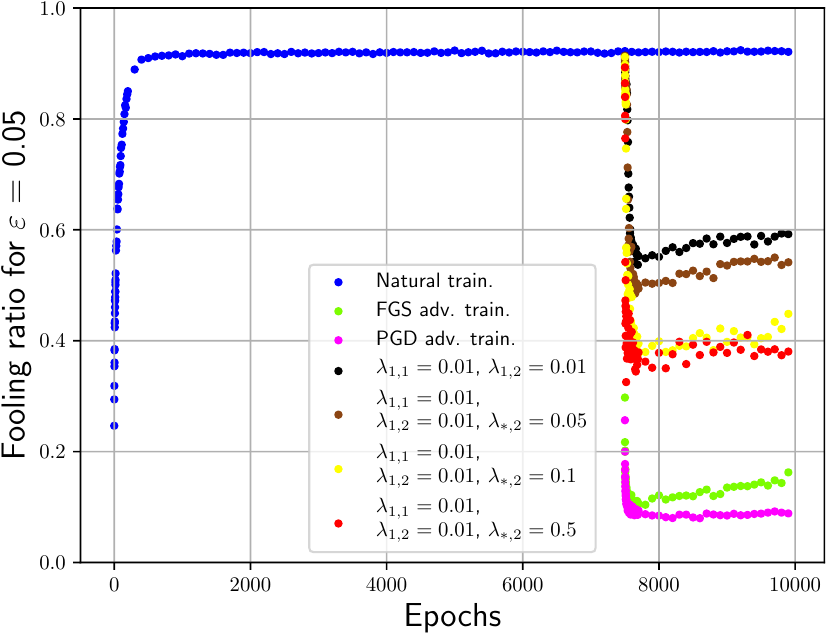}
            \centerline{\footnotesize{\textbf{(c)} \ac{PGD}}}\medskip
	\end{minipage} %
    \caption{Fooling ratios of the \ac{CNN} with respective attack and defense using a fixed $\epsilon = 0.05$. Dataset: F-MNIST.}
    \label{fig:fool_cnn}
\end{figure*}

\begin{figure*}[tbh]
   \centering
   \includegraphics[width=0.85\linewidth]{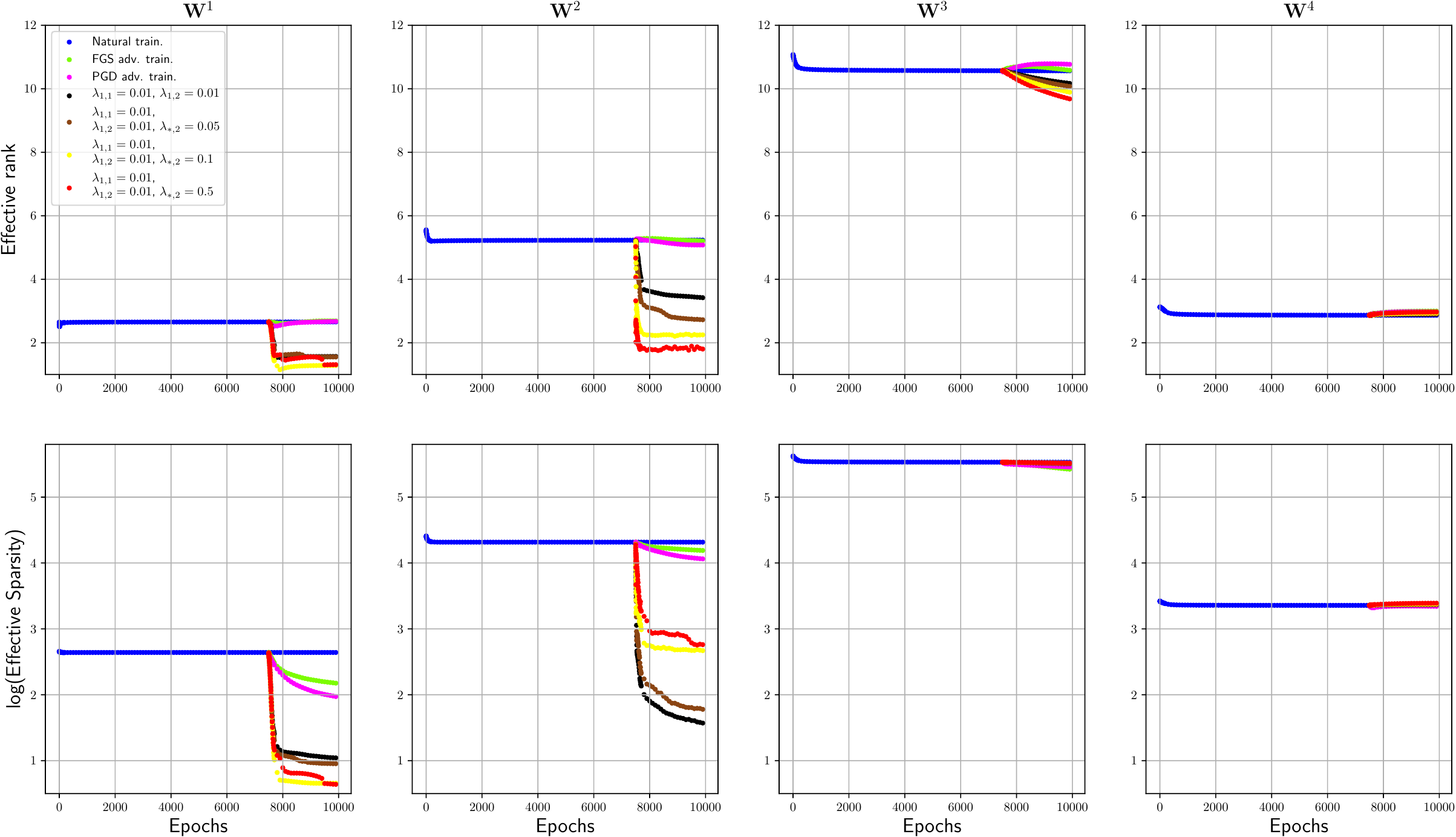}
   \caption{Effective rank and effective sparsity of the CNN.}
   \label{fig:compression_cnn}
\end{figure*}

\begin{figure}[tbh]
	\centering
    \begin{minipage}[b]{0.31\linewidth}
        	\includegraphics[width=0.98\linewidth]{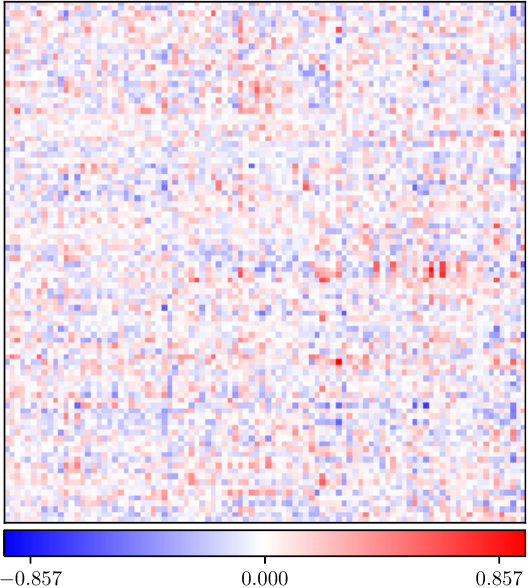}
            \centerline{\footnotesize{\textbf{(a)} Natural training}}\medskip
	\end{minipage} %
    \begin{minipage}[b]{0.31\linewidth}
        	\includegraphics[width=0.98\linewidth]{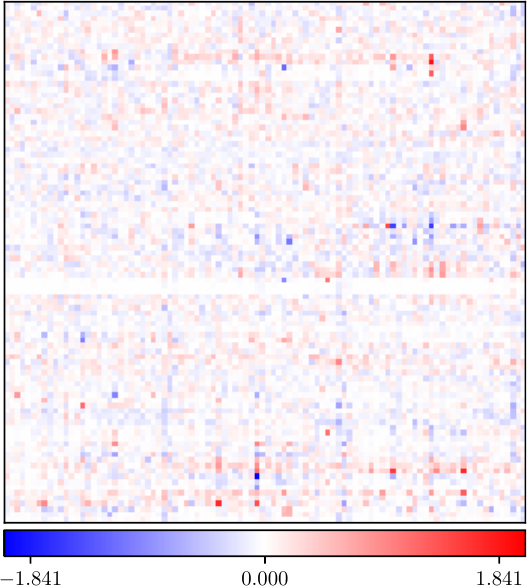}
            \centerline{\footnotesize{\textbf{(b)} \ac{FGSM} training}}\medskip
	\end{minipage} %
	\begin{minipage}[b]{0.31\linewidth}
        	\includegraphics[width=0.98\linewidth]{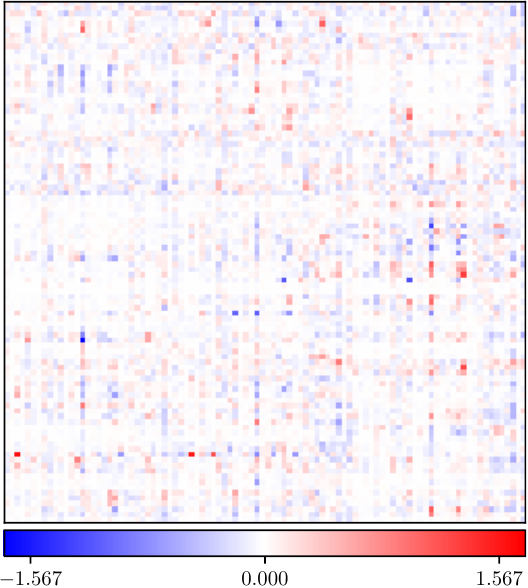}
            \centerline{\footnotesize{\textbf{(c)} \ac{PGD} training}}\medskip
	\end{minipage} %
    \caption{Reshaped second convolutional filter $\mat{W}^2 \in \mathbb{R}^{3 \times 3 \times 32 \times 32}$ of the \ac{CNN} after natural vs. adversarial training with $\epsilon  = 0.05$.}
    \label{fig:second_cnn_kernel_adv_train}
\end{figure}

\begin{figure}[tbh]
	\centering
	\begin{minipage}[b]{0.31\linewidth}
        	\includegraphics[width=0.98\linewidth]{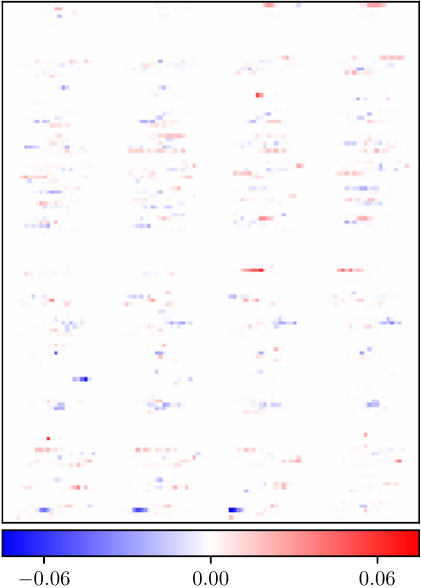}
	\end{minipage} %
	~
    \begin{minipage}[b]{0.31\linewidth}
        	\includegraphics[width=0.98\linewidth]{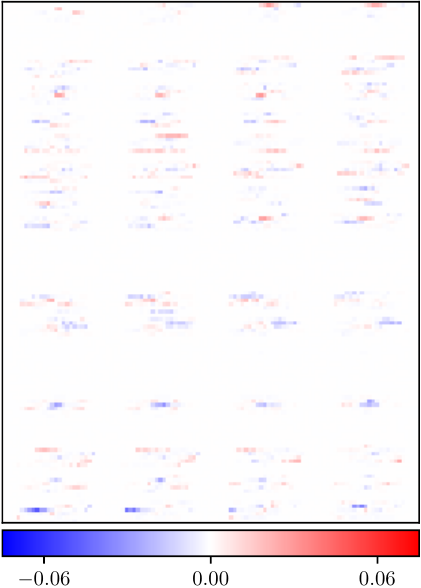}
	\end{minipage} %
	~
	\begin{minipage}[b]{0.31\linewidth}
        	\includegraphics[width=0.98\linewidth]{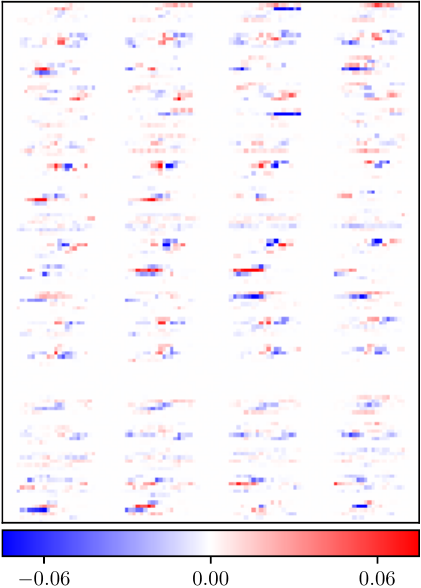}
	\end{minipage} %

	\begin{tabular}{ccc}
	%
		\\
        \textbf{(a)} ${\lambda}_{1, 1} = 0.01$ 
	&
        \textbf{(b)} ${\lambda}_{1, 1} = 0.01,$
	&
        \textbf{(c)} ${\lambda}_{1, 1} = 0.003,$
    \\
    &
    $\quad \lambda_{*, 1} = 0.01$
    &
    $\quad \lambda_{\text{joint}} = 0.04$
	\end{tabular}
    \caption{Reshaped input weight matrix $\mat{W}^1 \in \mathbb{R}^{20 \times 784}$ of the \ac{FCNN} after training with respective regularization.}
    \label{fig:first_fc_kernel_reg}
\end{figure}

\begin{figure}[tbh]
	\centering
	\begin{tabular}{cc}
    \begin{minipage}[b]{0.31\linewidth}
        	\includegraphics[width=0.98\linewidth]{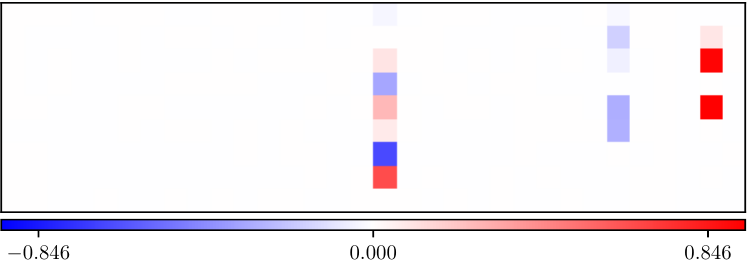}
	\end{minipage} %
		&
    \begin{minipage}[b]{0.31\linewidth}
        	\includegraphics[width=0.98\linewidth]{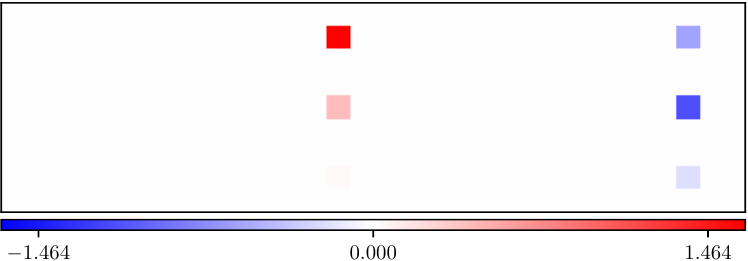}
	\end{minipage} %
		\\
        \textbf{(a)} ${\lambda}_{1, 1} = 0.01$, 
		&
        \textbf{(b)} ${\lambda}_{1, 1} = 0.01$, 
		\\
		$\quad {\lambda}_{1, 2} = 0.01$
		&
		$\quad \, \, {\lambda}_{1, 2} = 0.01$, 
		\\
		&
		$\, \, \, {\lambda}_{*, 2} = 0.5$
	\end{tabular}
    \caption{Reshaped input convolutional filter $\mat{W}^1 \in \mathbb{R}^{3 \times 3 \times 1 \times 32}$ of the \ac{CNN} after training with respective regularization.}
    \label{fig:first_cnn_kernel_reg}
\end{figure}

\begin{figure}[tbh]
	\centering
	\begin{tabular}{cc}
    \begin{minipage}[b]{0.31\linewidth}
        	\includegraphics[width=0.98\linewidth]{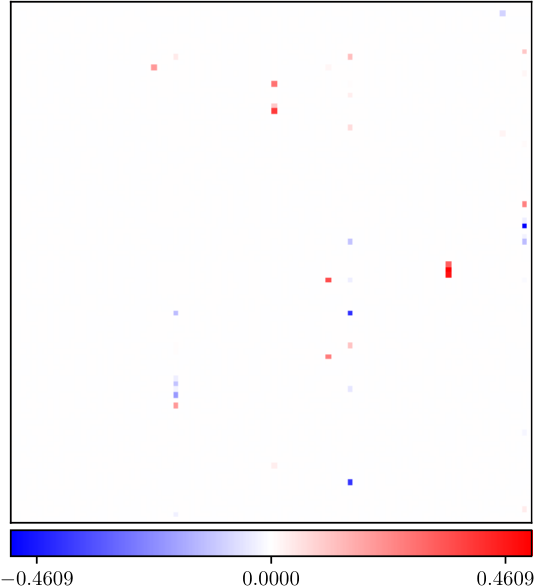}
	\end{minipage} %
		&
    \begin{minipage}[b]{0.31\linewidth}
        	\includegraphics[width=0.98\linewidth]{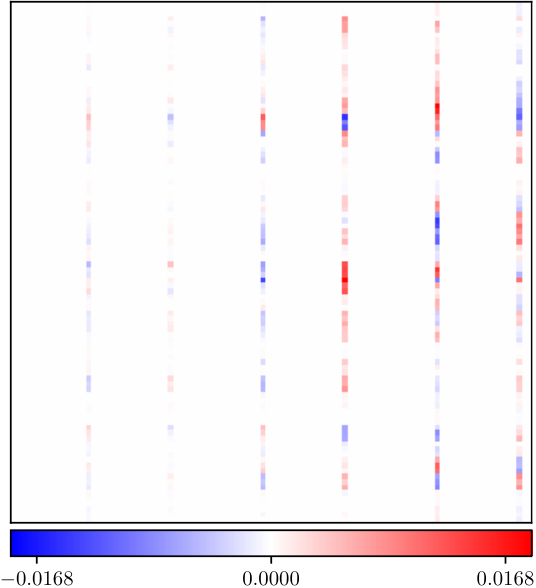}
	\end{minipage} %
		\\
        \textbf{(a)} ${\lambda}_{1, 1} = 0.01$, 
		&
        \textbf{(b)} ${\lambda}_{1, 1} = 0.01$, 
		\\
		$\quad {\lambda}_{1, 2} = 0.01$
		&
		$\quad \, \, {\lambda}_{1, 2} = 0.01$, 
		\\
		&
		$\, \, \, {\lambda}_{*, 2} = 0.5$
	\end{tabular}
    \caption{Reshaped second convolutional filter $\mat{W}^2 \in \mathbb{R}^{3 \times 3 \times 32 \times 32}$ of the \ac{CNN} after training with respective regularization.}
    \label{fig:second_cnn_kernel_reg}
\end{figure}

In the spirit of \cite{madry2018towards}, we raise the question to what happens to the weights $\mat{W}^i$ during adversarial training. 
For both settings, we took the naturally trained models and fine-tuned them adversarially with \ac{FGSM} and \ac{PGD} attacks using a fixed $\epsilon = 0.05$. 
As expected, adversarial training substantially decreases the fooling ratios. We illustrate the fooling ratios over the whole course of training in \figref{fig:fool_fc} and \figref{fig:fool_cnn} for the \ac{FCNN} and the \ac{CNN} respectively. We compute the fooling ratios for different types of attacks using the same fixed $\epsilon = 0.05$.
We also report the test accuracies on clean data (see the Appendix \ref{app:experiments}). For the \ac{CNN} model we observe slightly lower test accuracies for the adversarially trained models compared to the naturally trained one. 
In \figref{fig:compression_fc} and \figref{fig:compression_cnn}, we present the effective rank and the effective sparsity over the whole course of training for the \ac{FCNN} and the \ac{CNN} respectively. 
In case of the FCNN, we also estimate the mutual information between the input $\vec{x}$ and the hidden representations $\vec{t}^i$ using the KDE method \cite{KolchinskyT17, KolchinskyTW17} with a noise variance of ${\sigma}^2 = 0.1$. 
Note that the KDE method provides upper and lower bounds for mutual information, instead of a single estimate. 
In case of the FCNN, we find that adversarial training decreases the effective rank and the effective sparsity of some weight matrices as well as the mutual information (see \figref{fig:compression_fc}). 
This effect is most visible for the input weight matrix $\mat{W}^1 \in \mathbb{R}^{20 \times 784}$. This weight matrix is visualized for natural vs. \ac{FGSM}/\ac{PGD} adversarial training in \figref{fig:first_fc_kernel_adv_train}. 
While $\mat{W}^1$ looks like noise after natural training, it clearly looks like lower-rank and more sparse after \ac{FGSM}/\ac{PGD} adversarial training, with strongly correlated pattern visible.  
We can also observe this effect in an attenuated form for the CNN experiments. 
Further, \ac{FGSM}/\ac{PGD} adversarial training substantially decreases the effective sparsity of the first two convolutional filters $\mat{W}^1 \in \mathbb{R}^{3 \times 3 \times 1 \times 32}$ and $\mat{W}^2 \in \mathbb{R}^{3 \times 3 \times 32 \times 32}$ (see \figref{fig:compression_cnn}). 
Moreover, \ac{FGSM}/\ac{PGD} adversarial training slightly decreases the effective rank of $\mat{W}^2$ as well. 
Using \ac{PGD} adversarial training reduces the effective rank of $\mat{W}^2$ more than \ac{FGSM} adversarial training does as shown in \figref{fig:compression_cnn}. 
This is also visible in \figref{fig:second_cnn_kernel_adv_train} where we visualize $\mat{W}^2$ for natural vs. \ac{FGSM}/\ac{PGD} adversarial training. 
The weights after both \ac{FGSM} and \ac{PGD} adversarial training look more sparse and slightly lower-rank than after natural training. In addition, the weights after \ac{PGD} adversarial training look slightly lower-rank than after \ac{FGSM} adversarial training.
From these observations we first conclude that adversarial training leads to compression in the information theoretic sense. 
Then, we claim the following: 
\begin{claim}
\label{claim:adv_training}
Adversarial training leads to low-rank and sparse weights.
\end{claim}
%
\begin{figure}[tbh]
	\centering
    \begin{minipage}[b]{0.31\linewidth}
        	\includegraphics[width=0.98\linewidth]{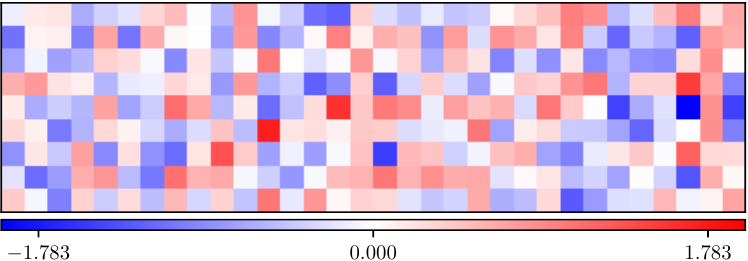}
            \centerline{\footnotesize{\textbf{(a)} Natural training}}\medskip
	\end{minipage} %
    \begin{minipage}[b]{0.31\linewidth}
        	\includegraphics[width=0.98\linewidth]{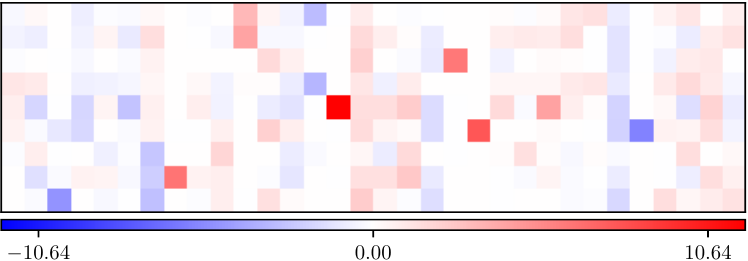}
            \centerline{\footnotesize{\textbf{(b)} \ac{FGSM} training}}\medskip
	\end{minipage} %
	\begin{minipage}[b]{0.31\linewidth}
        	\includegraphics[width=0.98\linewidth]{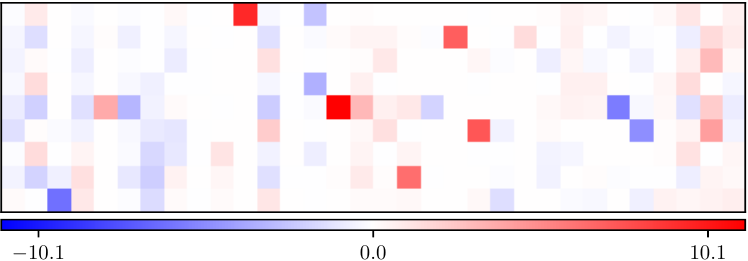}
            \centerline{\footnotesize{\textbf{(c)} \ac{PGD} training}}\medskip
	\end{minipage} %

    \caption{Reshaped input convolutional filter $\mat{W}^1 \in \mathbb{R}^{3 \times 3 \times 1 \times 32}$ of the \ac{CNN} after natural vs. adversarial training with $\epsilon = 0.05$.}
    \label{fig:first_cnn_kernel_adv_train}
    \end{figure}
%
%
Motivated by Claim \ref{claim:adv_training}, we mimic adversarial training by promoting low-rank and sparse weights. We are interested on observing the effect of low-rankness and sparsity on the robustness of \acp{DNN}. 
To that end we promote these properties through regularization of the loss function as explained in Section \ref{sec:inducing_compression_through_regularization}.
In case of the FCNN, we first only regularize the input weight matrix $\mat{W}^1$ with different choices of ${\lambda}_{1, 1}$ and 
${\lambda}_{*, 1}$ as this weight matrix substantially changes its effective rank and effective sparsity during \ac{FGSM}/\ac{PGD} adversarial training. 
We also provide results on performing only $l_1$-regularization on $\mat{W}^1$ with ${\lambda}_{1, 1} = 0.01$. 
Both kinds of regularization substantially decrease the fooling ratios (see \figref{fig:fool_fc}). 
As shown in \figref{fig:compression_fc}, 
both kinds of regularization substantially decrease the effective sparsity and effective rank of $\mat{W}^1$, as well as $\mat{W}^2$. Surprisingly, performing only $l_1$-regularization on $\mat{W}^1$ also decreases the effective rank of $\mat{W}^1$. 
We visualize $\mat{W}^1$ after training with both kinds of regularization in \figref{fig:first_fc_kernel_reg} indicating sparsity and low-rankness. 
As shown in \figref{fig:compression_fc}, we are able to simultaneously decrease the effective rank of $\mat{W}^1$, $\mat{W}^2$ and $\mat{W}^3$ by using the joint regularization technique \eqref{eq:joint}. Together with explicitly promoting sparsity on $\mat{W}^1$, we are able to again improve robustness compared to regularizing only $\mat{W}^1$.  
In case of the CNN, we regularize the reshaped versions of the first two convolutional filters $\mat{W}^1$, $\mat{W}^2$ (see Appendix \ref{app:reshape} for details about reshaping) with different choices of ${\lambda}_{1, 1}$, ${\lambda}_{1, 2}$ and ${\lambda}_{*, 2}$. We observe a significant improvement in terms of robustness (see \figref{fig:fool_cnn}) when increasing the explicit low-rank promotion on $\mat{W}^2$ by nuclear-norm regularization. In \figref{fig:compression_cnn}, we see that a considerably lower effective rank of $\mat{W}^2$ can be obtained by adding nuclear-norm regularization. This result further supports that simultaneous sparsity and low-rankness of the weights should be favored on the way towards adversarial robustness. 
In \figref{fig:first_cnn_kernel_reg} and \figref{fig:second_cnn_kernel_reg}, we visualize the first and second convolutional filter $\mat{W}^1$, $\mat{W}^2$ after training with the different kinds of regularization.
These findings support the second claim of this paper.
\begin{claim}
Simultaneously low-rank and sparse weights promote robustness against adversarial examples.
\end{claim}
%
%
%
%
\clearpage
\section{Discussion}
\label{sec:concl}
We are able to obtain more sparse and lower-rank weights by introducing new regularizations. Despite the improvement in adversarial robustness, we cannot still match the robustness of adversarial training.  This suggests that beside sparsity and low-rankness of the weights further attributes might be considered. Moreover, it raises an open question whether there is an optimal combination of sparsity and low-rankness of the weights that can further improve robustness. The experiments suggest that, despite their important role, sparsity and low-rankness of the weights do not fully explain the success of adversarial training.

%
\bibliographystyle{apacite}
\bibliography{arxiv-version}

\clearpage
\appendix

\section{Additional Experiments} \label{app:experiments}
%
%
Tables \ref{tab:test_acc_fcnn} and \ref{tab:test_acc_cnn} are self-explanatory.
\begin{table}[tbh]
    \centering
    \begin{tabular}{|l|l|l|l|l|}
    \hline
     \multicolumn{1}{|l|}{\textbf{Method}} & \begin{tabular}[l]{@{}l@{}} \textbf{Test accuracy} \\ \textbf{(on clean data)}\end{tabular} \\ \hline
    \multicolumn{1}{|l|}{Natural train.} &  $0.9508$  \\ \hline
     \ac{FGSM} adv. train. &  $0.9697$ \\ \hline
     \ac{PGD} adv. train. & $0.9689$ \\ \hline
     ${\lambda}_{1, 1} = 0.01$ & $0.9527$ \\ \hline
    ${\lambda}_{*, 1} = 0.01$ & $0.9681$ \\ \hline
     ${\lambda}_{1, 1} = 0.01$, ${\lambda}_{*, 1} = 0.01$ & $0.9446$ \\ \hline
     ${\lambda}_{1, 1} = 0.01$, ${\lambda}_{*, 1} = 0.03$ & $0.9327$ \\ \hline
    \end{tabular}
    \caption{Test accuracies FCNN. Dataset: MNIST.}
    \label{tab:test_acc_fcnn}
\end{table}
\begin{table}[tbh]
    \centering
    \begin{tabular}{|l|l|l|l|l|}
    \hline
    \multicolumn{1}{|l|}{\textbf{Method}} & \begin{tabular}[l]{@{}l@{}} \textbf{Test accuracy} \\ \textbf{(on clean data)}\end{tabular} \\ \hline
     Natural train. &  $0.9041$  \\ \hline
     \ac{FGSM} adv. train. &  $0.8639$ \\ \hline
     \ac{PGD} adv. train. & $0.864$ \\ \hline
     ${\lambda}_{1, 1} = 0.01$, ${\lambda}_{1, 2} = 0.01$, ${\lambda}_{*, 2} = 0$ & $0.8917$ \\ \hline
    ${\lambda}_{1, 1} = 0.01$, ${\lambda}_{1, 2} = 0.01$, ${\lambda}_{*, 2} = 0.05$ & $0.8623$ \\ \hline
     ${\lambda}_{1, 1} = 0.01$, ${\lambda}_{1, 2} = 0.01$, ${\lambda}_{*, 2} = 0.1$ & $0.8614$ \\ \hline
     ${\lambda}_{1, 1} = 0.01$, ${\lambda}_{1, 2} = 0.01$, ${\lambda}_{*, 2} = 0.5$ & $0.8399$ \\ \hline
    \end{tabular}
    \caption{Test accuracies CNN. Dataset: F-MNIST.}
    \label{tab:test_acc_cnn}
\end{table}

\section{Reshaping of Convolutional Filters} 
\label{app:reshape}

Assume a hidden representation composed of non-overlapping $3$-dimensional patches $\mat{x}_1, \dots, \mat{x}_m$.
A convolution operation is performed using $3$-dimensional weight tensors $\mat{W}_1, \dots, \mat{W}_n $ of the same size as the patches, which results in the following (vectorized) output of the convolution
\begin{align*}
&
\begin{pmatrix}
\mathrm{vec}(\mat W_1 )^{\T}\mathrm{vec}(\mat x_1 ) \\ 
\mathrm{vec}(\mat W_1 )^{\T}\mathrm{vec}(\mat x_2 ) \\
\vdots \\
\mathrm{vec}(\mat W_n )^{\T}\mathrm{vec}(\mat x_m )
\end{pmatrix}
\\
=& 
\underbrace{\left(
\underbrace{
\begin{pmatrix}
	\mathrm{vec}(\mat W_1 )^{\T} \\
	\mathrm{vec}(\mat W_2 )^{\T} \\
	\vdots \\
	\mathrm{vec}(\mat W_n )^{\T}
\end{pmatrix}}_{=: \widetilde{\mat W}}
\otimes \mat I_m
\right)}_{=: \mat W}
\begin{pmatrix}
	\mathrm{vec}(\mat x_1 ) \\
	\mathrm{vec}(\mat x_2 ) \\
	\dots \\
	\mathrm{vec}(\mat x_m )
\end{pmatrix}
\in \real^{mn} \, . 
\end{align*}
Therefore, this convolution operation can be written as a standard fully connected layer with weight matrix $\mat W = \widetilde{\mat W}\otimes \mat I_m$. Note that $\mat W$ and $\widetilde{\mat W}$ have the same effective rank, and $\widetilde{\mat W}$ is a reshaped version of the $4$-dimensional weight tensor composed of $W_1, \dots, W_n$.
Therefore, the matrix $\widetilde{\mat W}$ is used in \eqref{eq:nuc_reg} and \eqref{eq:joint} to regularize convolutional layers.

\end{document}